\documentclass[10pt,letterpaper,conference]{IEEEtran}

\usepackage{graphicx,times,amsmath,epstopdf} 
\usepackage{massieee}

\renewcommand{\arraystretch}{1.5}
\renewcommand{\IEEEQED}{\IEEEQEDopen}

\IEEEoverridecommandlockouts    

\begin{document}

\title{\ \\ \LARGE\bf A Fuzzy Directional Distance Measure
\thanks{$\dag$ Department of Computer Science, University of Nottingham,
Nottingham, Nottinghamshire, UK (email: psxjm5; christian.wagner; uwe.aickelin@nottingham.ac.uk).}
\thanks{$\ddag$ Department of Computer Science, Loughborough University,
Loughborough, Leicestershire, UK (email: C.J.Hinde@lboro.ac.uk).}
\thanks{This work was partially funded by the EPSRC’s Towards Data-Driven Environmental Policy Design grant, EP/K012479/1 and the RCUK’s Horizon Digital Economy Research Hub grant, EP/G065802/1.}}

\author{J.C. McCulloch$^\dag$, {\it Student Member, IEEE}, C.J. Hinde$^\ddag$, {\it Senior Member, IEEE},\\
	C. Wagner$^\dag$, {\it Senior Member, IEEE} and U. Aickelin$^\dag$}

\maketitle

\begin{abstract}
The measure of distance between two fuzzy sets is a fundamental tool within fuzzy set theory, however, distance measures currently within the literature use a crisp value to represent the distance between fuzzy sets. A real valued distance measure is developed into a fuzzy distance measure which better reflects the uncertainty inherent in fuzzy sets and a fuzzy directional distance measure is presented, which accounts for the direction of change between fuzzy sets. A multiplicative version is explored as a full maximal assignment is computationally intractable so an intermediate solution is offered.
\end{abstract}


\section{Introduction\label{sec:introduction}}
Distance measures for fuzzy sets (FSs) are an important tool and have been applied to many fields. There are many distance measures that are appropriate in different situations, for example Mahalanobis distance \cite{Mahalanobis:1936:DIST} was proposed in 1936 and many more are in use today, such as Chaudhur and Rosenfeld's \cite{Chaudhur:1996:Distance} distance measure for FSs and work by Dubois \cite{Dubois:1980:fuzzy}. An interesting application for many workers is case based reasoning, Segura et al. present a variety of case based distance measures \cite{Segura:2008:CBA}. 

While distance measures traditionally use a single real value to express distance, representing the distance as a FS would give a richer, more accurate comparison, reflecting the uncertainty inherent in FSs. This work follows and draws on work by \cite{McCulloch:2013:DDM}, which describes a real-valued directional distance measure, and presents a distance measure which describes distance as a FS. In \cite{McCulloch:2013:DDM}, alpha-cuts ($\alpha$-cuts) are used to measure distance by comparing each $\alpha$-cut of one FS with the same $\alpha$-cut of another FS. This, however, introduces difficulties for non-normal FSs where an $\alpha$-cut results in the empty set. Though the problem was addressed, the method taken is limited by using a substituted value of distance for $\alpha$-cuts where one of the fuzzy sets is not present. The method introduced in this paper removes this problem by comparing every $\alpha$-cut (or mass assignment) of one FS with every $\alpha$-cut of the other FS. This also results in a more accurate description of distance. This fuzzy distance measure is achieved using a mass assignment (MA) framework \cite{Baldwin:1984:FRIL,Baldwin:1995:FRIL}.

Section \ref{sec:background} provides a background on MAs and semantic unification of FSs which form the basis of the distance measure. Following this, Sections \ref{sec:background:distance} and \ref{sec:directional} introduce both a non-directional and directional distance measure, respectively. Demonstrations of the distance measure are then given for non-normal and non-convex fuzzy FSs in Sections \ref{sec:non normal} and \ref{sec:multimodal}, respectively.
Finally, Section \ref{sec:conclusions} presents some conclusions.

\section{Background\label{sec:background}}
A background on MA and semantic unification is presented first. Mass assignment uses a measure of support based on semantic unification, \cite{Baldwin:1984:FRIL} that is generalised in \cite{Hinde:2007:RAI} and further in \cite{Hinde:2008:SUSS}. Distance is commonly calculated using $\alpha$-cuts, which are related to MAs, such that they both break down the FS along the membership axis. The crucial difference is that, using $\alpha$-cuts, the membership of an element is ascertained by the maximum $\alpha$-cut to which it belongs, whereas using masses the membership value is given by the sum of the masses. As the two methods are related, the MA techniques should be applicable to a distance measure just as they are applicable to a support measure. 

\subsection{Mass Assignments \label{sec:background:mass}}
Mass is a precise amount of probability assigned to a set of events, rather than individual events. A MA defined on the domain $C$ is written as \cite{Baldwin:1991:TMA}:

 \begin{equation}
 \begin{array}{l}
 X = X_1:x_1, ...., X_N:x_n \\
 \text{where } \sum^n_{i=1} x_i = 1.0 \\
 \text{and } X_i \in 2^C 
 \end{array}
 \end{equation}
where $2^C$ is the powerset of the domain $C$, $X_i$ is a subset of the domain $C$, and $x_i$  is the amount of mass assigned to $X_i$. For example, consider the FS $F$ expressed as
\begin{equation}
  \nonumber
  F = \{x, \mu_F(x)\ |\ x \in X \}
\end{equation}
 where $X$ is the discrete space
\begin{equation}
  \nonumber
  X = \{x_1, x_2,.... x_n\}.
\end{equation}
To calculate the mass of $F$, its elements are first ordered such that \cite{Baldwin:1991:TMA}
\begin{equation}
  \nonumber
  \mu_F(x_i) \geq \mu_F(x_j) \text{ if } i < j
\end{equation}
The MA of the FS $F$ is then calculated as follows \cite{Baldwin:1991:TMA,Baldwin:1994:Algebra}
\begin{equation}
  \begin{array}{l}
    m_F = \{ \{x_1,..., x_i\}: \mu_F(x_i) - \mu_F(x_{i+1}),\\ 
    \ \ \ \ \ \ \ \emptyset :1-\mu(x_1) \}  \text{   with } \mu_F(x_{n+1})=0
 \end{array}
 \label{eqn:mass_assignment}
\end{equation}
Note that if the FS is normalised then the mass assigned to the empty set will be 0.
For example, given two FSs $A$ and $G$ \\
\begin{eqnarray*}
A &=& \set{1.0 \mid a,\ 0.7 \mid b,\ 0.2 \mid c} \\
G &=& \set{0.9 \mid a,\ 0.6 \mid b,\ 0.1 \mid c}
\end{eqnarray*}
with set of support $\set{a,b,c}$, the masses assigned to $A$ and $G$ are $m_A\set{A_i:a_i}$ and $m_G\set{G_i:g_i}$ as follows
\begin{eqnarray*}
m_A &=& \set{a}:0.3,\set{a,b}:0.5, \set{a,b,c}:0.2 \\
m_G &=& \set{a}:0.3,\set{a,b}:0.5, \set{a,b,c}:0.1, \emptyset:0.1
\end{eqnarray*}

Having briefly covered MAs of FSs, the next section introduces semantic unification which will be the basis of the distance measure in this paper.

\subsection{Semantic Unification\label{sec:background:unification}}
Semantic unification assesses the support of a claim $A$ given a ground clause $G$. As defined in \cite{Baldwin:1995:FRIL}, it does not deal with claims or ground evidence that are inconsistent. However, an extended version described in \cite{Hinde:2007:RAI} deals with inconsistent FSs, or alternatively non-normalised FSs. This work starts with the extended version which is defined as follows for two MAs $m_A = \set{A_i: a_i}$ and $m_G = \set{G_j: g_j}$:
\begin{equation}
  \scriptstyle{T(A_i\mid G_j) =}
    \begin{cases}
      \set{t} & \text{if } A_i \supseteq G_j \vee ( A_i = \emptyset \wedge G_j = \emptyset)\\
      \set{f} & \text{if } A_i \cap G_j = \emptyset \wedge A_i \neq \emptyset\wedge G_j \neq \emptyset\\
      \emptyset & \text{if} A_i \neq \emptyset \wedge G_j = \emptyset\\
      \set{f,t} & \text{otherwise}
    \end{cases}
    \label{eq:unification}
\end{equation}
This can be read as
\begin{itemize}
 \item The truth of A is true if G supports A
 \item The truth of A is false if G denies A
 \item The truth of A is unknown if G is unknown (no evidence exists)
 \item The truth of A is inconsistent if G both supports and denies A
\end{itemize}
An example of semantic unification using the MAs of $A$ and $G$ detailed above is given in Table \ref{tab:unification:example}. The calculations multiply the masses of the contributing sets to calculate the mass of the resulting set. By adding the final masses assigned to each set the result $A\mid G = \set{t}:0.67,\set{f,t}:0.23,\emptyset:0.1$ is obtained.

\begin{table}[!htbp]
  \caption{Semantic unification operator incorporating inconsistency for non-normalised FSs and assigning mass multiplicatively.\label{tab:unification:example}}
  \setlength{\tabcolsep}{8pt}
  \renewcommand{\arraystretch}{1.2}
  \begin{tabular}{c c|c|c|c|c|}
    & \multicolumn{1}{c}{} & \multicolumn{4}{c}{G} \\
    & \multicolumn{1}{c}{$A\mid G$} & \multicolumn{1}{c}{$\set{a}:$} & \multicolumn{1}{c}{$\set{a,b}:$} & 
	  \multicolumn{1}{c}{$\set{a,b,c}:$} & \multicolumn{1}{c}{$\emptyset:$} \\
    & \multicolumn{1}{c}{} & \multicolumn{1}{c}{0.3} & \multicolumn{1}{c}{0.5} & 
	  \multicolumn{1}{c}{0.1} & \multicolumn{1}{c}{0.1} \\ \cline{3-6}
    & $\set{a}:$ & $\set{t}:$ & $\set{f,t}:$ & $\set{f,t}:$ & $\emptyset:$ \\ 
    & 0.3 & 0.09 & 0.15 & 0.03 & 0.03 \\ \cline{3-6}
    A & $\set{a,b}:$ & $\set{t}:$ & $\set{t}:$ & $\set{f,t}:$ & $\emptyset:$ \\
    & 0.5 & 0.15 & 0.25 & 0.05 & 0.05 \\ \cline{3-6}
    & $\set{a,b,c}:$ & $\set{t}:$ & $\set{t}:$ & $\set{t}:$ & $\emptyset:$ \\
    & 0.2 & 0.06 & 0.1 & 0.02 & 0.02 \\ \cline{3-6}
  \end{tabular}
\end{table}

Semantic unification thus delivers a FS of truth values indicating the degree of support the fuzzy claim $A$ receives from the fuzzy evidence $G$. Neither FS is necessarily normalised and so the FS representing the degree of support, similarly, is not necessarily normalised. Though the calculations above multiply the masses to calculate the mass of the resulting set, this is not the most general answer possible. For example, Table \ref{tab:unification:maximal} shows a possible maximal assignments applied to the FSs $A$ and $G$. Calculating the maximal assignment involves maximising the value assigned to $\set{f,t}$, then maximising either to $\set{f}$ or $\set{t}$, and then finally assigning mass to $\emptyset$. Maximising first to $\set{f,t}$, then $\set{t}$, $\set{f}$ and $\emptyset$ results in $A\mid G = \set{t}:0.5,\set{f,t}:0.4,\emptyset:0.1$, more uncertain than the multiplicative result.

\begin{table}[!htbp]
  \caption{Semantic unification operator incorporating inconsistency for non-normalised FSs and assigning mass maximally.\label{tab:unification:maximal}}
  \setlength{\tabcolsep}{7pt}
  \renewcommand{\arraystretch}{1.2}
  \begin{tabular}{c c|c|c|c|c|}
    & \multicolumn{1}{c}{} & \multicolumn{4}{c}{G} \\
    & \multicolumn{1}{c}{$A\mid G$} & \multicolumn{1}{c}{$\set{a}:$} & \multicolumn{1}{c}{$\set{a,b}:$} & 
	  \multicolumn{1}{c}{$\set{a,b,c}:$} & \multicolumn{1}{c}{$\emptyset:$} \\
    & \multicolumn{1}{c}{} & \multicolumn{1}{c}{0.3} & \multicolumn{1}{c}{0.5} & 
	  \multicolumn{1}{c}{0.1} & \multicolumn{1}{c}{0.1} \\ \cline{3-6}
    & $\set{a}:$ & $\set{t}:$ & $\set{f,t}:$ & $\set{f,t}:$ & $\emptyset:$ \\ 
    & 0.3 & 0.0 & 0.3 & 0.0 & 0.0 \\ \cline{3-6}
    A & $\set{a,b}:$ & $\set{t}:$ & $\set{t}:$ & $\set{f,t}:$ & $\emptyset:$ \\
    & 0.5 & 0.3 & 0.1 & 0.1 & 0.0 \\ \cline{3-6}
    & $\set{a,b,c}:$ & $\set{t}:$ & $\set{t}:$ & $\set{t}:$ & $\emptyset:$ \\
    & 0.2 & 0.0 & 0.1 & 0.0 & 0.1 \\ \cline{3-6}
  \end{tabular}
\end{table}

\section{Distance Measures\label{sec:background:distance}}
In \cite{McCulloch:2013:DDM} a distance measure is based on measuring distance between individual $\alpha$-cuts. As discussed earlier, MA is also based on $\alpha$-cuts and so the generalisation is straightforward as presented next. For the distance measure proposed in this paper it is difficult to obtain a maximal MA in the general case, and even in the case analysed here a full maximal assignment is not available; however, a better approximation than the multiplicative case is presented.

\subsection{Mass based distance measure\label{sec:distance}}
The MA operator based on a non-directional Hausdorff distance measure \cite{Chaudhur:1996:Distance, McCulloch:2013:DDM} is given in (\ref{eqn:distance}). Using MAs, the two intervals $a_{i,l}$ and $b_{j,k}$ are sets of possibilities, such that $a_{i,l}$ represents all points in $A_i$ and $b_{j,k}$ represents all points in $B_j$. To calculate the distance between two subsets $A_i$ and $B_j$ the following equation is used:

\begin{equation}
  D(A_i,B_j)=
  \begin{cases}
    \{ \vert a_{i,l} - b_{j,k} \vert \} & if A_i \neq \emptyset \wedge B_j \neq \emptyset \\
    \emptyset & otherwise 
  \end{cases}
  \label{eqn:distance}
\end{equation}

The non-directional distance measure is altered into a directional distance measure \cite{McCulloch:2013:DDM} as follows (\ref{eqn:dirdistance}):

\begin{equation}
  D(A_i,B_j) =
  \begin{cases}
    \{ b_{j,k} - a_{i,l} \} & if A_i \neq \emptyset \wedge B_j \neq \emptyset \\
    \emptyset & otherwise 
  \end{cases}
  \label{eqn:dirdistance}
\end{equation}

Note that the operation has been reversed from $a_{i,l} - b_{j,k}$ in (\ref{eqn:distance}) to $b_{j,k} - a_{i,l}$ within (\ref{eqn:dirdistance}), and the absolute value of the distance is no longer used. This is to account for the directional nature of the distance measure, and results in MAs assigned to the positive domain where the FS $B$ is placed to the right of $A$ within the universe of discourse, and MAs in the negative domain otherwise.

\begin{figure}[!htbp]
  \begin{center}
    \includegraphics[scale=0.23]{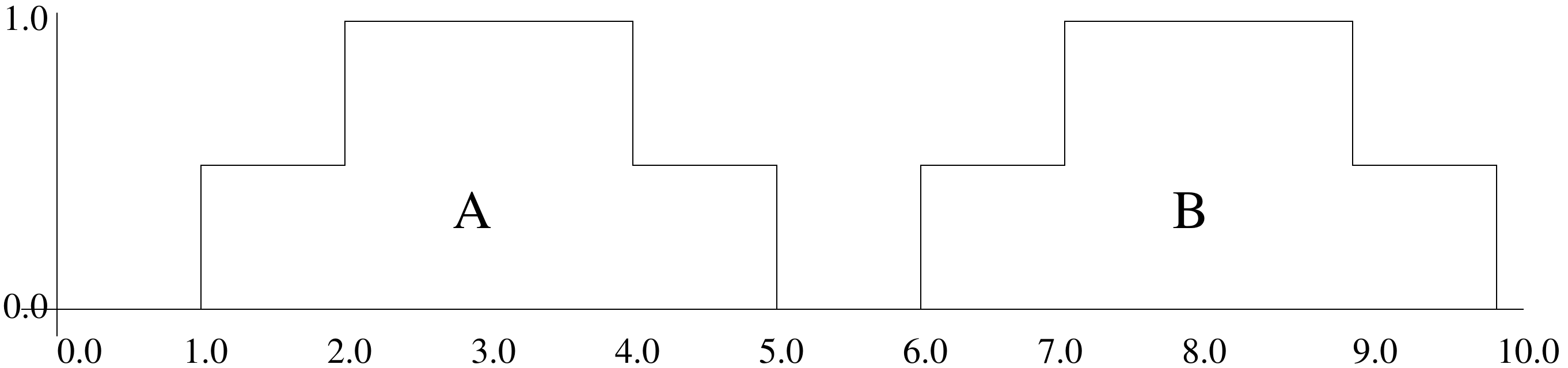}
    \caption{Fuzzy sets $A$ and $B$.\label{fig:aandb}}
  \end{center}
\end{figure}

Table \ref{tab:distance:matrix} shows the calculation of the non-directional distance (\ref{eqn:distance}) between the two sets $A$ and $B$ as shown in Fig. \ref{fig:aandb}. For simplicity, $A$ and $B$ are two highly discretised fuzzy numbers. The MAs of $A$ and $B$ using (\ref{eqn:mass_assignment}), are $m_A$ and $m_B$ as follows:

\begin{eqnarray*}
m_A &=& [1.0,5.0]:0.5, [2.0,4.0]:0.5\\
m_B &=& [6.0,10.0]:0.5, [7.0,9.0]:0.5
\end{eqnarray*}

To derive the distance between $A$ and $B$, the distance measure given in (\ref{eqn:distance}) is used and the masses are multiplied as shown in Table \ref{tab:distance:matrix}.

\begin{table}[h!]
  \caption{Distance measure between $A$ and $B$ assigned multiplicatively}
  \begin{tabular}{c c|c|c|}
    & \multicolumn{1}{c}{} & \multicolumn{2}{c}{$B$}\\
    & \multicolumn{1}{c}{$D(A, B)$} & \multicolumn{1}{c}{[6.0,10.0]:} & \multicolumn{1}{c}{[7.0,9.0]:}\\
    & \multicolumn{1}{c}{} & \multicolumn{1}{c}{0.5} & \multicolumn{1}{c}{0.5}\\ \cline{3-4}
    & [1.0,5.0]: & [1.0,9.0]: & [2.0,8.0]:\\
    $A$ & 0.5 & 0.25 & 0.25 \\ \cline{3-4}
    & [2.0,4.0]: & [2.0,8.0]: & [3.0,7.0]: \\
    & 0.5 & 0.25 & 0.25 \\ \cline{3-4}
\end{tabular}
\label{tab:distance:matrix}
\end{table}

From Table \ref{tab:distance:matrix}, the following MAs and corresponding FS are obtained
\begin{eqnarray}
\nonumber
m_{D(A,B)}&=&[1.0,9.0]:0.25,[2.0,8.0]:0.5,\\
&&[3.0,7.0]:0.25\label{eqn:prodmass}\\
\nonumber
D(A,B)&=&\{0.25\mid [1.0,2.0], 0.5\mid [2.0,3.0],\\
\nonumber
&&1.0 \mid [3.0,7.0], 0.5 \mid [7.0,8.0],\\
&&0.25 \mid [8.0,9.0]\}\label{eqn:prodfuzz}
\end{eqnarray}

\begin{figure}[!htbp]
  \begin{center}
    \includegraphics[scale=0.23]{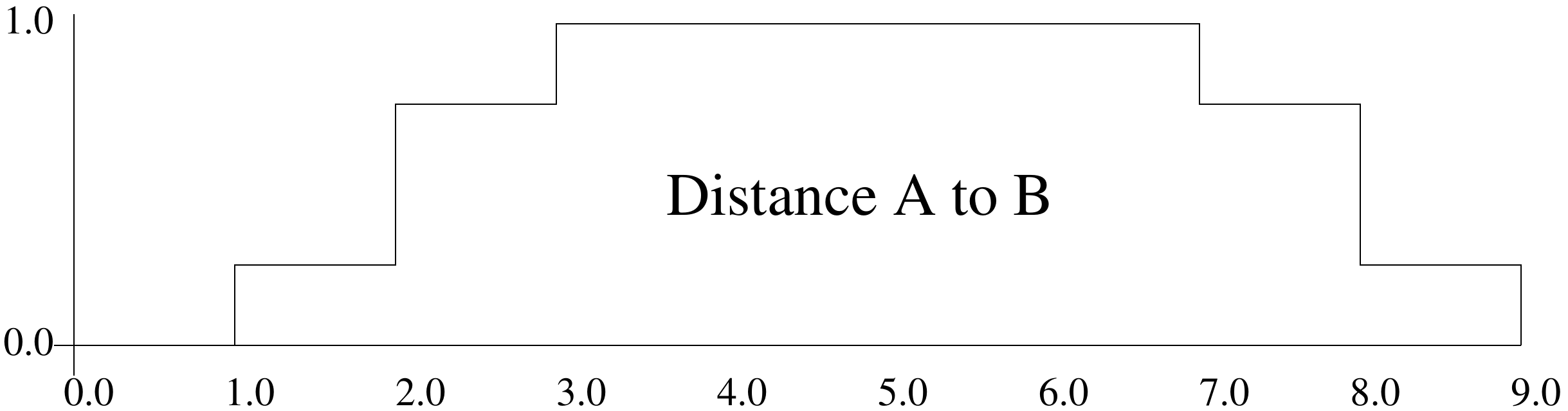}
    \caption{Distance between FSs $A$ and $B$ shown as a FS.\label{fig:distanceAB}}
  \end{center}
\end{figure}

Fig. \ref{fig:distanceAB} shows the FS representing the distance between $A$ and $B$ with the masses assigned multiplicatively. Note that the smallest distance between any two points of $A$ and $B$ is 1 and the largest is 9, both of which are conveyed in the end points of the FS in Fig. \ref{fig:distanceAB}.

A distance measure between FSs has been introduced using multiplicative MA, distance using maximal MAs is addressed next .

\subsection{Maximal assignments\label{sec:maximal}}
The definition of a maximal assignment is one that cannot be reached by means of restrictions or linear combination of any of the other possible assignments. The two types of restriction of concern are defined in (\ref{eqn:type1}), Type 1, and (\ref{eqn:type2}), Type 2.

\begin{equation}
\small
m' = \left\{ \begin{array}{c}
\set{L_i:m_i}\cup\set{L_j:m_j + x} \cup \set{L_k:m_k - x}\\
L_k \supseteq L_j, x \leq m_k, i \neq j, i \neq k\\
\end{array}
\right\}\label{eqn:type1}
\end{equation}
 
\begin{equation}
\small
m' = \left\{ \begin{array}{c}
\set{L_i:m_i}\cup \set{L_k:m_k - x} \cup \\
\set{L_n:m_n - x}\cup \set{L_u:m_u + x}\\
\cup \set{L_p:m_p + x}\mid\\
L_u = L_k \cup L_n, L_p = L_k \cap L_n,\\
L_i \neq L_k,L_n,L_u,L_p
\end{array}
\right\}\label{eqn:type2}
\end{equation}
 
The distance measures are special cases of MAs. If both numbers are triangular FSs then the final result is also a triangular FS.

\begin{theorem}
If the two FSs are similar isosceles triangles then all entries in a distance measure assignment matrix are subsets, supersets or equal to one another.
\begin{IEEEproof}
Let the two triangles $A$ and $B$ be defined by the parameters as below, then the intervals will be of the form:\\
$[(B_l+B_n{\delta}) - (A_u-A_n{\delta}),(B_u-B_n{\delta}) - (A_l+A_n{\delta})]$\\
which may be rewritten as\\
$[(B_l-A_u+(B_n{\delta} +A_n{\delta}),(B_u-A_l -(B_n{\delta} +A_n{\delta})]$\\
where $A_l$ and $A_u$, $B_l$ and $B_u$ are the lower bound and upper bound points of the triangles $A$ and $B$, respectively;\\
$A_n$ and $B_n$ are the heights of the slices measured in number of slices;\\
${\delta}$ is the amount the side of the triangle increases with each slice.

The rates of change of each quantity in the intervals are identical so the result follows immediately. Once $\delta$ has been chosen the lower and upper bounds of the intervals are fixed. 
\end{IEEEproof}
\end{theorem}
\begin{corollary}
Theorem 1 essentially means there are no type 2 restrictions for similar isosceles triangles.
\end{corollary}

\begin{theorem}
If the two base FSs are not similar isosceles triangles then there may be entries in a distance measure assignment matrix that are overlapping intervals and are not subsets.
\begin{IEEEproof}
  Let the two triangles be defined by the parameters as below, then the intervals will be of the form:
  \begin{eqnarray}
      [(B_l+B_n{\delta_L}) - (A_u-A_n{\delta_R}), \nonumber \\
      (B_u-B_n{\delta_R}) - (A_l+A_n{\delta_L})] \nonumber
  \end{eqnarray}
  where $A_n$ and $B_n$ are the heights of the slices measured in number of slices;\\
  ${\delta_L}$ is the amount the left hand side of the triangle increases with each slice;\\
  ${\delta_R}$ is the amount the right hand side of the triangle decreases with each slice.

  The two quantities of interest from above are:\\
  $(B_n{\delta_L}) + A_n{\delta_R}),(B_n{\delta_R} + A_n{\delta_L})$ \\
  If ${\delta_L} > {\delta_R}$ then if $B_n$ is reduced by 1 and $A_n$ increased by one, $(B_n{\delta_L} + A_n{\delta_R})$ will be reduced while simultaneously $(B_n{\delta_R} + A_n{\delta_L})$ will be raised. Thus, the two intervals (\ref{eqn:first_interval}) and (\ref{eqn:second_interval}) overlap.
  \begin{eqnarray}
    \nonumber
    [(B_l+B_n{\delta_L}) - (A_h-A_n{\delta_R}),\\
    (B_h-B_n{\delta_R}) - (A_l+A_n{\delta_L})]\label{eqn:first_interval}
  \end{eqnarray}
  \begin{eqnarray}
    \nonumber
    [(B_l+(B_n-1){\delta_L}) - (A_h-(A_n+1){\delta_R}),\\
    (B_h-(B_n-1){\delta_R}) - (A_l+(A_n+1){\delta_L})]\label{eqn:second_interval}
  \end{eqnarray}
\end{IEEEproof}
\end{theorem}

\begin{corollary}
Theorem 2 essentially means there may be type 2 restrictions if the two triangles are not similar and isosceles.
\end{corollary}

The theorems above show that type 2 restrictions are likely to occur in many situations. If only type 1 restrictions are considered then it is easy to see that the distance measure between two nested FSs lies down the main diagonal. This is computationally straightforward and results in more general assignments than the multiplicative assignment. An assumption of independence between the two sets is now not necessary.

Taking this approach with unification, the mass is maximally assigned along the diagonal, as shown in Table \ref{tab:distance:matrix:max} which measures the FSs $A$ and $B$ in Fig. \ref{fig:aandb}.
\begin{table}[!htbp]
  \caption{Distance measure between $A$ and $B$ assigned maximally}
  \begin{tabular}{c c|c|c|}
    & \multicolumn{1}{c}{} & \multicolumn{2}{c}{$B$}\\
    & \multicolumn{1}{c}{$D(A, B)$} & \multicolumn{1}{c}{[6.0,10.0]:} & \multicolumn{1}{c}{[7.0,9.0]:}\\
    & \multicolumn{1}{c}{} & \multicolumn{1}{c}{0.5} & \multicolumn{1}{c}{0.5}\\ \cline{3-4}
    & [1.0,5.0]: & [1.0,9.0]: & [2.0,8.0]:\\
    $A$ & 0.5 & 0.5 & 0.0 \\ \cline{3-4}
    & [2.0,4.0]: & [2.0,8.0]: & [3.0,7.0]: \\
    & 0.5 & 0.0 & 0.5 \\ \cline{3-4}
\end{tabular}
\label{tab:distance:matrix:max}
\end{table}

Resulting in
\begin{eqnarray}
m_{D(A,B)}&=&[1.0,9.0]:0.5,[3.0,7.0]:0.5\label{eqn:diagmass}\\
\nonumber
D(A,B)&=&\{0.5\mid [1.0,3.0], 1.0\mid [3.0,7.0],\\
&&0.5 \mid [7.0,9.0]\}\label{eqn:diagfuzz}
\end{eqnarray}

\begin{figure}[!htbp]
  \begin{center}
    \includegraphics[scale=0.23]{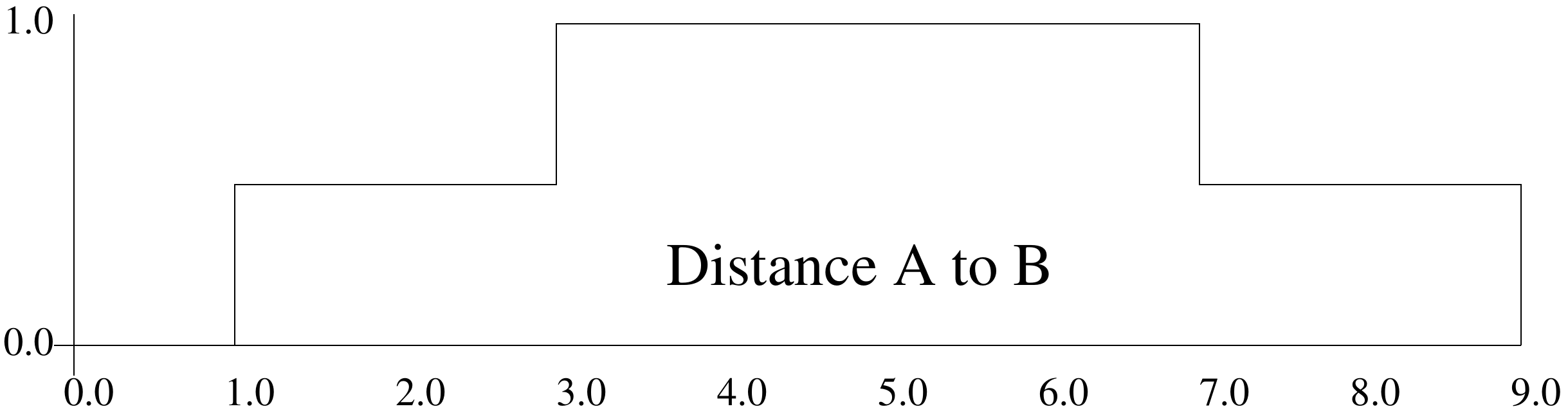}
    \caption{Distance between FSs $A$ and $B$ obtained down the diagonal shown as a FS.\label{fig:distanceABMax}}
  \end{center}
\end{figure}

Fig. \ref{fig:distanceABMax} shows the FS representing the distance between $A$ and $B$ with the masses assigned down the diagonal.

The assignment in Fig. \ref{fig:distanceABMax} should be restrictable to the assignment shown in Fig. \ref{fig:distanceAB} using type 1 or type 2 restrictions, however neither assignment is reachable from the other. Alternatively, taking the assignment from the other diagonal gives Table \ref{tab:distance:matrix:revdiag} resulting in the assignment in (\ref{eqn:revdiagmass}) and (\ref{eqn:revdiagfuzz}).

\begin{table}[!htbp]
  \caption{Distance measure between $A$ and $B$ assigned down the other diagonal, downwards and right to left}
  \begin{tabular}{c c|c|c|}
    & \multicolumn{1}{c}{} & \multicolumn{2}{c}{$B$}\\
    & \multicolumn{1}{c}{$D(A, B)$} & \multicolumn{1}{c}{[6.0,10.0]:} & \multicolumn{1}{c}{[7.0,9.0]:}\\
    & \multicolumn{1}{c}{} & \multicolumn{1}{c}{0.5} & \multicolumn{1}{c}{0.5}\\ \cline{3-4}
    & [1.0,5.0]: & [1.0,9.0]: & [2.0,8.0]:\\
    $A$ & 0.5 & 0.0 & 0.5 \\ \cline{3-4}
    & [2.0,4.0]: & [2.0,8.0]: & [3.0,7.0]: \\
    & 0.5 & 0.5 & 0.0 \\ \cline{3-4}
\end{tabular}
\label{tab:distance:matrix:revdiag}
\end{table}

Resulting in
\begin{eqnarray}
m_{D(A,B)}&=&[2.0,8.0]:1.0\label{eqn:revdiagmass}\\
D(A,B)&=&\{1.0\mid [2.0,8.0]\}\label{eqn:revdiagfuzz}
\end{eqnarray}

\begin{figure}[!htbp]
  \begin{center}
    \includegraphics[scale=0.23]{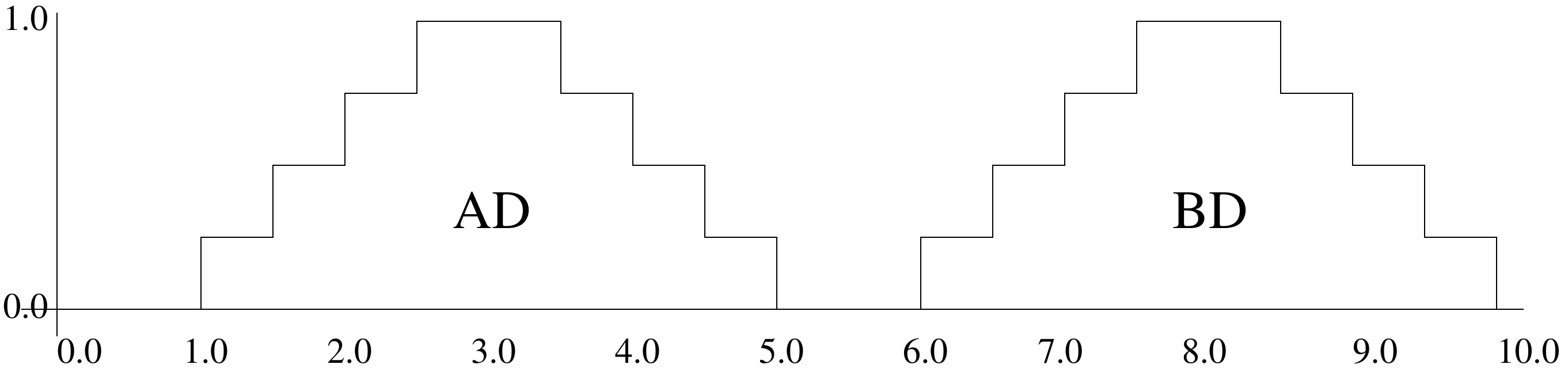}
    \caption{Fuzzy sets $AD$ and $BD$.\label{fig:adandbd}}
  \end{center}
\end{figure}

Given the two assignments (\ref{eqn:diagmass}) and (\ref{eqn:revdiagmass}), a linear combination results in the multiplication assignment (\ref{eqn:prodmass}). There are no type 2 restrictions and the two orthogonal assignments, when linearly combined, result in the product assignment. At this point it is unclear that this is a reasonable assumption. Yet, consider a more detailed view of $A$ and $B$, as $AD$ and $BD$, shown in Fig. \ref{fig:adandbd}


The MAs of $AD$ and $BD$, denoted $m_{AD}$ and $m_{BD}$, are
\begin{eqnarray*}
m_{AD} &=& [1.0,5.0]:0.25, [1.5,4.5]:0.25,\\
&&[2.0,4.0]:0.25, [2.5,3.5]:0.25\\
m_{BD} &=& [6.0,10.0]:0.25, [6.5,9.5]:0.25,\\
&&[7.0,9.0]:0.25, [7.5,8.5]:0.25
\end{eqnarray*}
To derive the distance between $AD$ and $BD$, the distance measure given in (\ref{eqn:distance}) is used and the masses are multiplied resulting in (\ref{eqn:prodmassadbd}) and (\ref{eqn:prodfuzzadbd}).

{\arraycolsep=1.2pt\def\arraystretch{1}
\begin{equation}
  \begin{array}{l l}
    {\scriptstyle m_{D(AD,BD)}}=&[1.0,9.0]:0.0625,[1.5,8.5]:0.125,\\
    & [2.0,8.0]:0.1875,[2.5,7.5]:0.25,\\
    & [3.0,7.0]:0.1875,[3.5,6.5]:0.125,\\
    & [4.0,6.0]:0.0625
  \end{array}
  \label{eqn:prodmassadbd}
\end{equation}
}

{\arraycolsep=1.2pt\def\arraystretch{1}
\begin{equation}
  \begin{array}{l l}
    D(AD,BD)=&\{0.25\mid [1.0,2.0], 0.5\mid [2.0,3.0],\\
    & 1.0 \mid [3.0,7.0], 0.5 \mid [7.0,8.0],\\
    & 0.25 \mid [8.0,9.0]\}
  \end{array}
  \label{eqn:prodfuzzadbd}
\end{equation}
}

The assignment down the left to right diagonal is shown in Table \ref{tab:distanceadbd:maxmatrix}.

\begin{table}[h!]
  \caption{Distance measure between $AD$ and $BD$ assigned down the left to right diagonal}
\setlength{\tabcolsep}{8.5pt}
\tiny
  \begin{tabular}{c c|c|c|c|c|}
    & \multicolumn{1}{c}{} & \multicolumn{4}{c}{$BD$}\\
    & \multicolumn{1}{c}{D($AD$, $BD$)} & \multicolumn{1}{c}{[6.0,10.0]:} & \multicolumn{1}{c}{[6.5,9.5]:}& \multicolumn{1}{c}{[7.0,9.0]:} & \multicolumn{1}{c}{[7.5,8.5]:}\\
    & \multicolumn{1}{c}{} & \multicolumn{1}{c}{0.25} & \multicolumn{1}{c}{0.25}\\ \cline{3-6}
    & [1.0,5.0]: & [1.0,9.0]: & [1.5,8.5]:& [2.0,8.0]: & [2.5,7.5]:\\
    & 0.25 & 0.25 & 0.0 & 0.0 & 0.0 \\ \cline{3-6}
    & [1.5,4.5]: & [1.5,8.5]: & [2.0,8.0]: & [2.5,7.5]: & [3.0,7.0]: \\
    & 0.25 & 0.0 & 0.25 & 0.0 & 0.0 \\ \cline{3-6}
    $AD$ & [2.0,4.0]: & [2.0,8.0]: & [2.5,7.5]:& [3.0,7.0]: & [3.5,6.5]:\\
    & 0.25 & 0.0 & 0.0 & 0.25 & 0.0 \\ \cline{3-6}
    & [2.5,4.5]: & [2.5,7.5]: & [3.0,7.0]: & [3.5,6.5]: & [4.0,6.0]: \\
    & 0.25 & 0.0 & 0.0 & 0.0 & 0.25 \\ \cline{3-6}
\end{tabular}
\label{tab:distanceadbd:maxmatrix}
\end{table}

From Table \ref{tab:distanceadbd:maxmatrix} the following MAs and corresponding FS are obtained
{\arraycolsep=1pt\def\arraystretch{1}
\begin{equation}
 \begin{array}{l l}
  m_{D(AD,BD)}=&[1.0,9.0]:0.25,[2.0,8.0]:0.25,\\
  &[3.0,7.0]:0.25,[4.0,6.0]:0.25
  \label{eqn:maxmassadbd}
 \end{array}
\end{equation}
}

{\arraycolsep=1pt\def\arraystretch{1}
\begin{equation}
 \begin{array}{l l}
  D(AD,BD) =&\{0.25\mid [1.0,2.0], 0.5\mid [2.0,3.0],\\
  & 1.0 \mid [3.0,7.0], 0.5 \mid [7.0,8.0],\\
  & 0.25 \mid [8.0,9.0]\}
  \label{eqn:maxfuzzadbd}
 \end{array}
\end{equation}
}

Again this does not restrict to the product assignment and other orthogonal assignments are needed to make it possible to create a linear combination resulting in the product assignment. The crucial point to see is that four orthogonal assignments are needed to complete the process. The number required rises with the number of slices taken, and the computation quickly becomes intractable.

The diagonal assignment is one of the maximal orthogonal set and the assumption now is that taking this is a better solution than the product MA when applied to the distance measure. It should be noted that this assumption may not be justified for a general MA operator.

\section{Directional distance\label{sec:directional}}
This section describes the directional version of the distance operator (\ref{eqn:dirdistance}) with examples of the calculation.

Referring to the two FSs $A$ and $B$ in Fig. \ref{fig:aandb}, the calculation matrix shown in Table \ref{tab:dirdistance:matrixab} is the distance between $A$ and $B$ using the directional distance measure (\ref{eqn:dirdistance}). The masses of $A$ and $B$ are as follows
\begin{eqnarray*}
m_A &=& [1.0,4.0]:0.5, [2.0,3.0]:0.5\\
m_B &=& [6.0,9.0]:0.5, [7.0,8.0]:0.5
\end{eqnarray*}
Calculating the distance using (\ref{eqn:dirdistance}), as shown in Table \ref{tab:dirdistance:matrixab}, results in the  following MAs and FS.
\begin{eqnarray*}
  m_{D(A,B)}&=&[2.0,8.0]:0.5,[4.0,6.0]:0.5\\
  D(A,B)&=&\{0.5\mid [2.0,4.0], 1.0 \mid [4.0,6.0], \\
  && 0.5 \mid [6.0,8.0]\}
\end{eqnarray*}

\begin{table}[!htbp]
  \caption{The directional distance measure between FSs $A$ and $B$ calculated maximally.\label{tab:dirdistance:matrixab}}
  \begin{tabular}{c c|c|c|}
    & \multicolumn{1}{c}{} & \multicolumn{2}{c}{$B$}\\
    & \multicolumn{1}{c}{$D(A, B)$} & \multicolumn{1}{c}{[6.0,9.0]:} & \multicolumn{1}{c}{[7.0,8.0]:} \\
    & \multicolumn{1}{c}{} & \multicolumn{1}{c}{0.5} & \multicolumn{1}{c}{0.5} \\ \cline{3-4}
    & [1.0,4.0]: & [2.0,8.0]: & [3.0,7.0]: \\
    $A$ & 0.5 & 0.5 & 0.0 \\ \cline{3-4}
    & [2.0,3.0]: & [3.0,7.0]: & [4.0,6.0]: \\
    & 0.5 & 0.0 & 0.5 \\ \cline{3-4}
  \end{tabular}
\end{table}

Using the same FSs, and thus the same MAs, if the calculation is reversed to measure the distance from $B$ to $A$, as shown in Table \ref{tab:dirdistance:matrixba}, the quantities are now reversed and the numbers are negative in comparison to Table \ref{tab:dirdistance:matrixab}.

\begin{table}[!htbp]
  \caption{Directional distance measure between FSs $B$ and $A$.\label{tab:dirdistance:matrixba}}
  \begin{tabular}{c c|c|c|}
    & \multicolumn{1}{c}{} & \multicolumn{2}{c}{$A$}\\
    & \multicolumn{1}{c}{$D(B, A)$} & \multicolumn{1}{c}{[1.0,4.0]:} & \multicolumn{1}{c}{[2.0,3.0]:} \\
    & \multicolumn{1}{c}{} & \multicolumn{1}{c}{0.5} & \multicolumn{1}{c}{0.5} \\ \cline{3-4}
    & [6.0,9.0]: & [-2.0,-8.0]: & [-3.0,-7.0]: \\
    $B$ & 0.5 & 0.5 & 0.0  \\ \cline{3-4}
    & [7.0,8.0]: & [-3.0,-7.0]: & [-4.0,-6.0]: \\
    & 0.5 & 0.0 & 0.5  \\ \cline{3-4}
  \end{tabular}
\end{table}

\begin{figure}[!htbp]
  \begin{center}
    \includegraphics[scale=0.23]{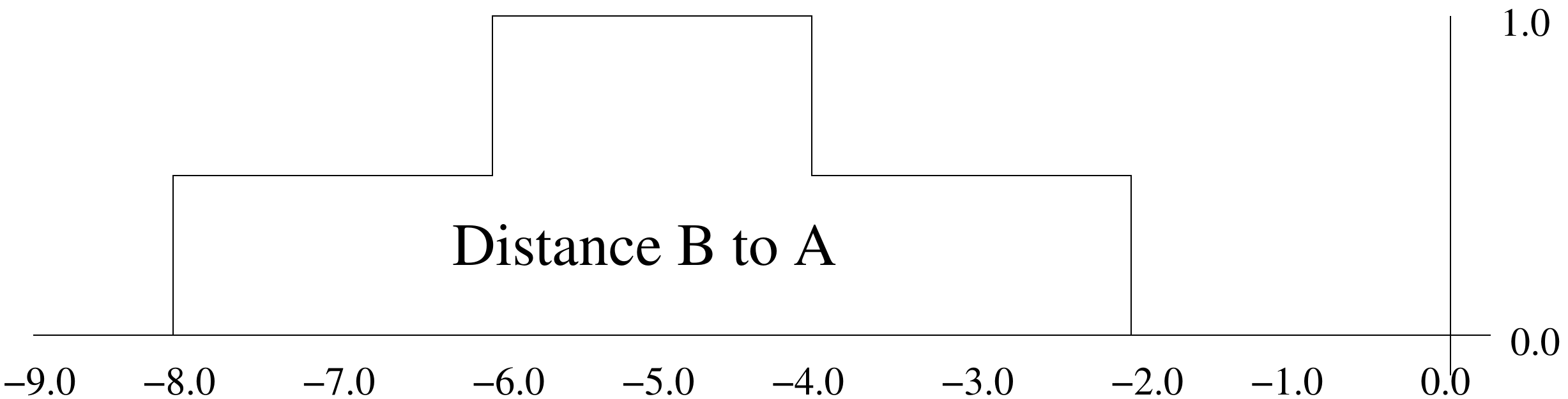}
    \caption{Distance between FSs $B$ and $A$ shown as a FS.\label{fig:distancebanda}}
  \end{center}
\end{figure}

The resulting FS is as follows and shown in Fig. \ref{fig:distancebanda}.
\begin{eqnarray*}
  m_{D(B,A)}&=&[-2.0,-8.0]:0.5,[-4.0,-6.0]:0.5\\
  D(B,A)&=&\{0.5\mid [-2.0,-4.0],1.0 \mid [-4.0,-6.0],\\
&&0.5 \mid [-6.0,-8.0]\}
\end{eqnarray*}

\section{Non-normal FSs and distances\label{sec:non normal}}
This section shows the effect of non-normal FSs in the distance calculation. 

\subsection{General case of non-normal distance}
\begin{figure}[!htbp]
  \begin{center}
    \includegraphics[scale=0.23]{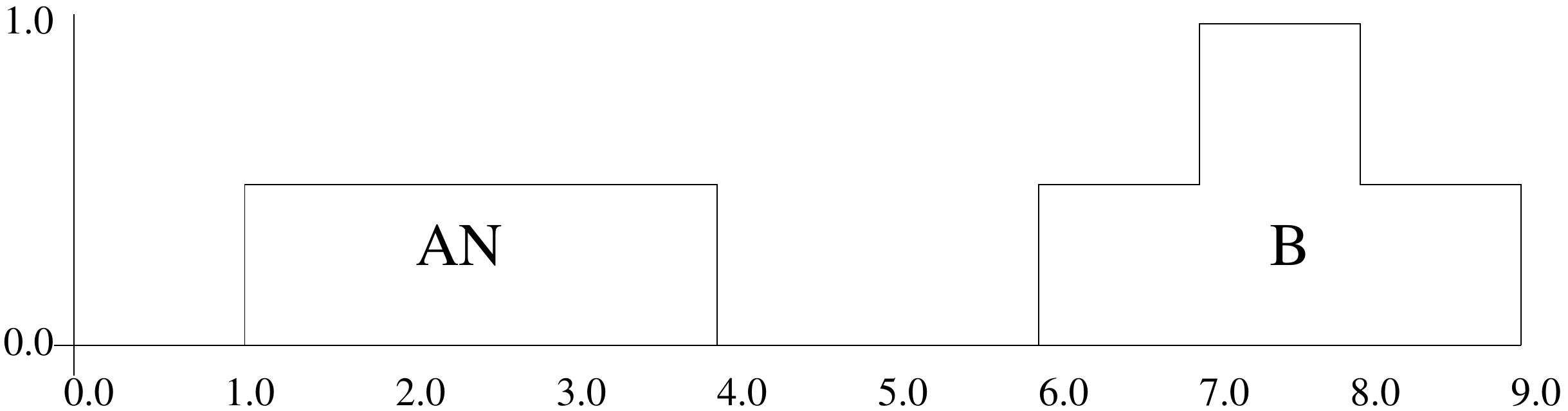}
    \caption{FSs $AN$ and $B$.\label{fig:aandbnon}}
  \end{center}
\end{figure}

The definition in (\ref{eqn:distance}) caters for non-normalised sets so redefining $A$ as $AN$ in Fig. \ref{fig:aandbnon} results in the calculation shown in Table \ref{tab:distancenon:matrix}. In this case the maximal assignment has to take care of assignment to the empty set. The maximal assignment would be to assign the mass along the diagonal, in this case the product assignment is more satisfactory but requires the assumption of independence to be justified.

\vspace{-0.5cm}

\begin{eqnarray*}
m_{AN} &=& [1.0,4.0]:0.5, []:0.5\\
m_B &=& [6.0,9.0]:0.5, [7.0,8.0]:0.5]
\end{eqnarray*}

\begin{table}[!htbp]
  \caption{Multiplicative distance measure between FSs $AN$ and $B$.\label{tab:distancenon:matrix}}
  \begin{tabular}{c c|c|c|}
    & \multicolumn{1}{c}{} & \multicolumn{2}{c}{$B$}\\
    & \multicolumn{1}{c}{$D(AN, B)$} & \multicolumn{1}{c}{[6.0,9.0]:} & \multicolumn{1}{c}{[7.0,8.0]:} \\
    & \multicolumn{1}{c}{} & \multicolumn{1}{c}{0.5} & \multicolumn{1}{c}{0.5} \\ \cline{3-4}
    & [1.0,4.0]: & [2.0,8.0]: & [3.0,7.0]: \\
    $AN$ & 0.5 & 0.25 & 0.25 \\ \cline{3-4}
    & []: & []: & []: \\
    & 0.5 & 0.25 & 0.25 \\ \cline{3-4}
  \end{tabular}
\end{table}

Table \ref{tab:distancenon:matrix} results in
{\arraycolsep=1pt\def\arraystretch{1.4}
\begin{equation}
 \nonumber
 \begin{array}{l l}
  m_{D(AN,B)} = & [2.0,8.0]:0.5,[3.0,7.0]:0.25, []:0.5\\
  D(AN,B) = & \{0.25\mid [2.0,3.0], 0.5\mid [3.0,7.0],\\
  & 0.25\mid [7.0,8.0]\}
 \end{array}
\end{equation}
}

\begin{figure}[!htbp]
  \begin{center}
    \includegraphics[scale=0.23]{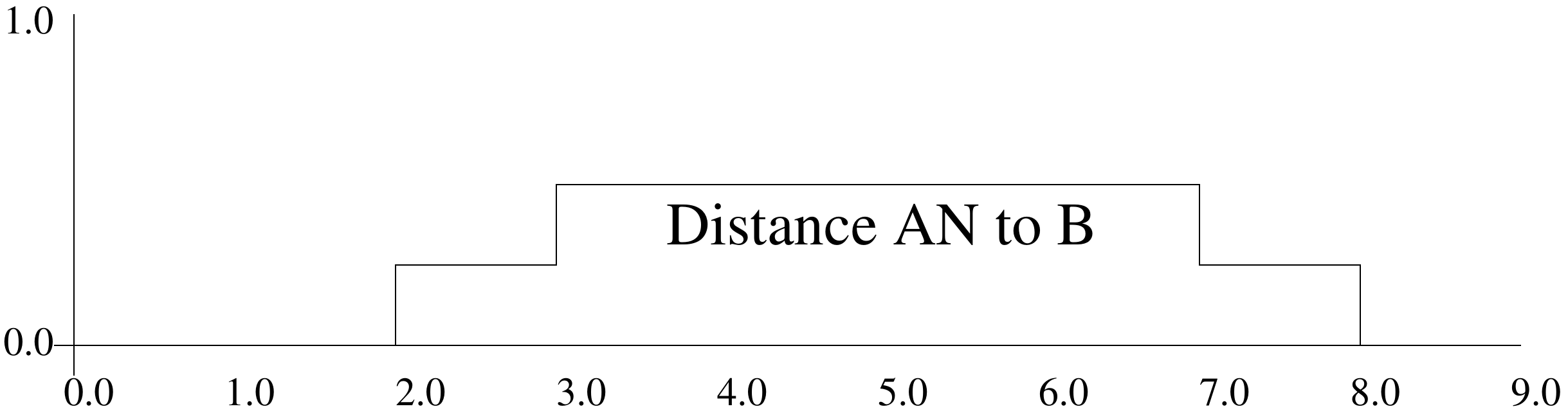}
    \caption{Distance between FSs $AN$ and $B$ shown as a FS.\label{fig:distanceaandbnon}}
  \end{center}
\end{figure}

The product fuzzy distance between $AN$ and $B$ is shown pictorially in Fig. \ref{fig:distanceaandbnon}.

The last example, in Fig. \ref{fig:distanceaandbnon}, shows that the distance between numbers where non-normalisation is involved is itself a non-normalised FS. It makes sense that it is not possible to measure the distance between sets that don't exist, and so the distance measure itself does not exist either. 



\subsection{Maximum likelihood estimates\label{sec:distance:maximum likelihood}}
All the distributions used may be transformed to a single interval by taking the maximum likelihood, least prejudiced values \cite{Baldwin:1995:FRIL} and performing the operations on the transformed values.

Transforming FSs $A$ and $B$ in Fig. \ref{fig:aandb} to maximum likelihood values results in $A = [2.0,3.0]$ and $B = [7.0,8.0]$, and the distance between $A$ and $B$ comes out at $D(A,B) = [4.0,6.0]$, which accords with standard interval arithmetic. It should be noted here that taking the centres of gravity yields a slightly different answer, with $A = 2.5$, $B = 7.5$ and $D(A,B) = 5.0$, which is more precise and also inaccurate as the uncertainty is not preserved. This will become much clearer when multimodal FSs are dealt with in Section \ref{sec:multimodal}. 

\section{Multimodal distance of non-convex sets\label{sec:multimodal}}
\begin{figure}[!htbp]
  \begin{center}
    \includegraphics[scale=0.23]{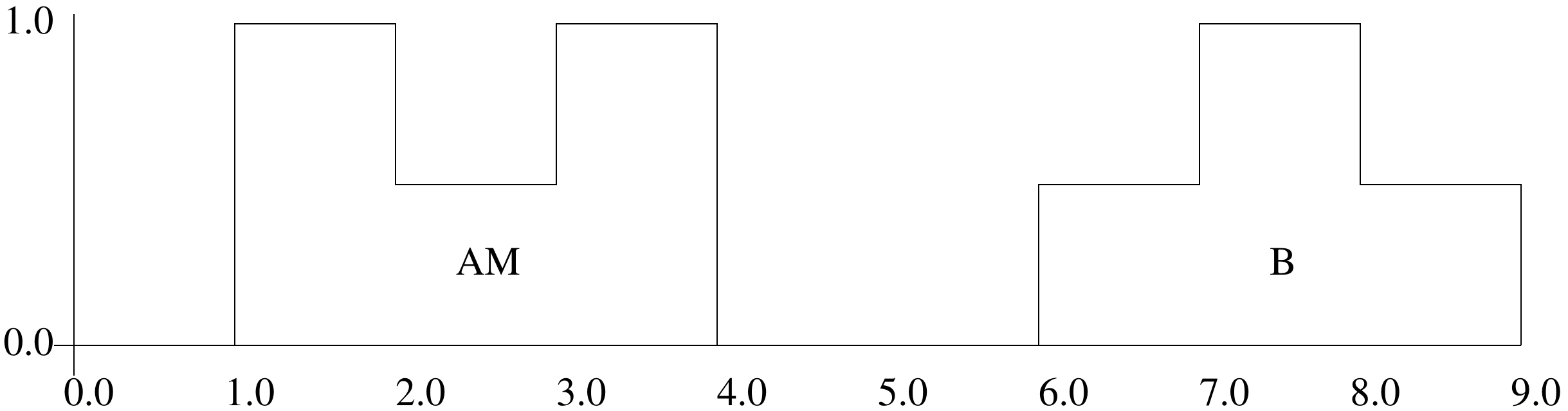}
    \caption{The multimodal FS $AM$ and $B$.\label{fig:amandb}}
  \end{center}
\end{figure}

This section describes the effect of calculating the distance between a bimodal FS and a unimodal FS. The distance between these FSs should intuitively be bimodal, but that is not necessarily the case. For example, take the two FSs $AM$ and $B$, shown in Fig. \ref{fig:amandb}:
\begin{eqnarray*}
m_{AM} &=& [1.0,4.0]:0.5, [1.0,2.0],[3.0,4.0]:0.5\\
m_{B} &=& [6.0,9.0]:0.5, [7.0,8.0]:0.5
\end{eqnarray*}

\begin{table}[!htbp]
  \caption{Distance measure between FSs $AM$ and $B$.\label{tab:distance:matrixAmB}} 
  \begin{tabular}{c c|c|c|}
    & \multicolumn{1}{c}{} & \multicolumn{2}{c}{$B$}\\
    & \multicolumn{1}{c}{$D(AM, B)$} & \multicolumn{1}{c}{[6.0,9.0]:} & \multicolumn{1}{c}{[7.0,8.0],} \\
    & \multicolumn{1}{c}{} & \multicolumn{1}{c}{0.5} & \multicolumn{1}{c}{0.5} \\ \cline{3-4}
    & [1.0,4.0]: & [2.0,8.0]: & [3.0,7.0]: \\
    $AM$ & 0.5 & 0.5 & 0.0 \\ \cline{3-4}
    & {[1.0,2.0],} & [4.0,8.0], & [5.0,7.0], \\
    & {[3.0,4.0]:} & [2.0,6.0]: & [3.0,5.0]: \\
    & 0.5 & 0.0 & 0.5 \\ \cline{3-4}
  \end{tabular}
\end{table}

The distance is calculated in Table \ref{tab:distance:matrixAmB}, resulting in
\begin{eqnarray*}
m_{D(AM,B)}&=&[2.0,8.0]:0.5,[3.0,7.0]:0.5\\
D(AM,B)&=&\{0.5\mid [2.0,3.0], 1.0\mid [3.0,7.0],\\
&&0.5 \mid [7.0,8.0]\}
\end{eqnarray*}

\begin{figure}[!htbp]
  \begin{center}
    \includegraphics[scale=0.23]{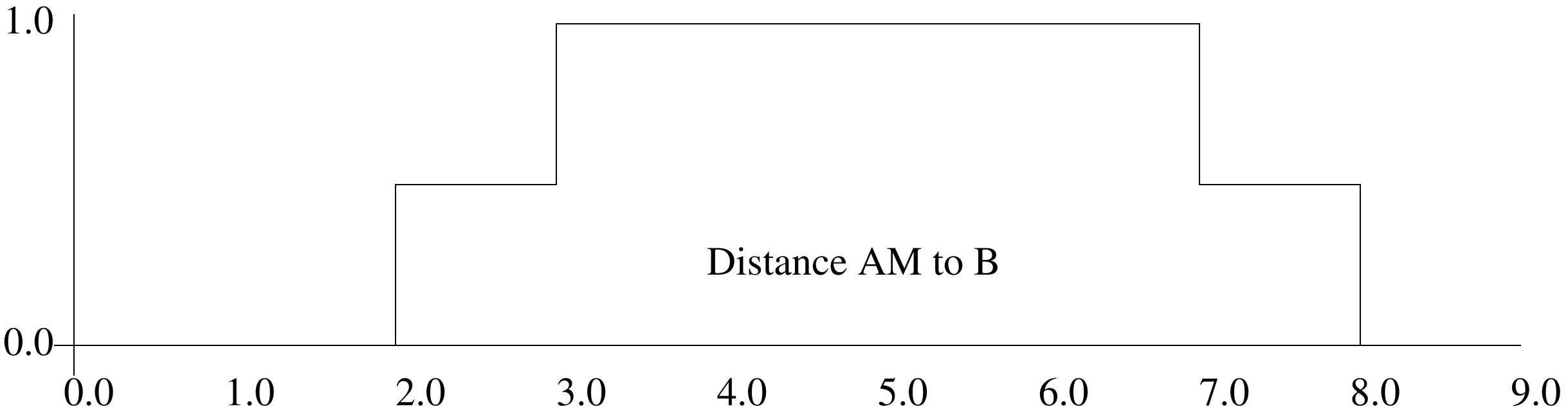}
    \caption{The unimodal FS showing the distance between the multimodal FS $AM$ and $B$.\label{fig:distanceamandb}}
  \end{center}
\end{figure}

\begin{figure}[!htbp]
  \begin{center}
    \includegraphics[scale=0.23]{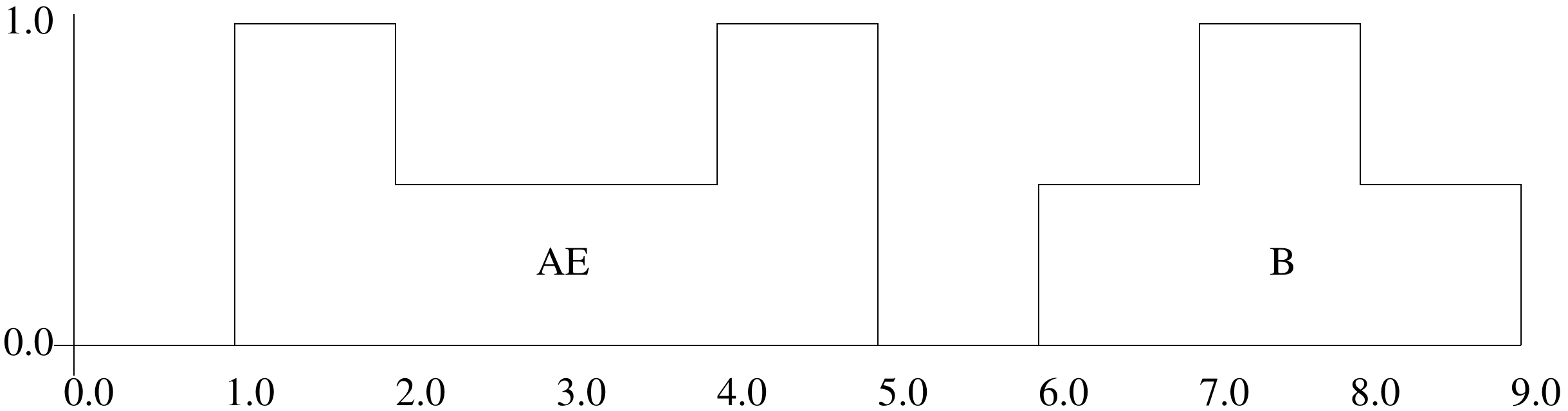}
    \caption{The multimodal FS $AE$ ($AM$ extended) and $B$.\label{fig:aeandb}}
  \end{center}
\end{figure}

The measure between $AM$ and $B$, shown in Fig. \ref{fig:distanceamandb}, results in a wider FS than the measure between $A$ and $B$, shown in Fig. \ref{fig:distancebanda}, because of the uncertainty about $AM$. However, it is not multimodal. Extending the width of $AM$ to $AE$ (see Fig. \ref{fig:aeandb}) results in a bimodal distance measure.

\begin{eqnarray*}
m_{AE} &=& [1.0,5.0]:0.5, [1.0,2.0],[4.0,5.0]:0.5\\
m_{B} &=& [6.0,9.0]:0.5, [7.0,8.0]:0.5
\end{eqnarray*}

\begin{table}[!htbp]
  \caption{Example distance measure between FSs $AE$ and $B$.\label{fig:distance:matrixAeB}}
  \begin{tabular}{c c|c|c|}
    & \multicolumn{1}{c}{} & \multicolumn{2}{c}{$B$}\\
    & \multicolumn{1}{c}{$D(AE, B)$} & \multicolumn{1}{c}{[6.0,9.0]:} & \multicolumn{1}{c}{[7.0,8.0],} \\
    & \multicolumn{1}{c}{} & \multicolumn{1}{c}{0.5} & \multicolumn{1}{c}{0.5} \\ \cline{3-4}
    & [1.0,5.0]: & [1.0,8.0]: & [2.0,7.0]: \\
    $AE$ & 0.5 & 0.5 & 0.0 \\ \cline{3-4}
    & {[1.0,2.0],} & [4.0,8.0], & [5.0,7.0], \\
    & {[4.0,5.0]:} & [1.0,5.0]: & [2.0,4.0]: \\
    & 0.5 & 0.0 & 0.5 \\ \cline{3-4}
  \end{tabular}
\end{table}

\begin{table}[!htbp]
  \caption{Example distance measure between FSs $AE$ and $B$, simplified.\label{fig:distance:matrixAeBSimp}}
  \begin{tabular}{c c|c|c|}
    & \multicolumn{1}{c}{} & \multicolumn{2}{c}{$B$}\\
    & \multicolumn{1}{c}{$D(AE, B)$} & \multicolumn{1}{c}{[6.0,9.0]:} & \multicolumn{1}{c}{[7.0,8.0],} \\
    & \multicolumn{1}{c}{} & \multicolumn{1}{c}{0.5} & \multicolumn{1}{c}{0.5} \\ \cline{3-4}
    & [1.0,5.0]: & [1.0,8.0]: & [2.0,7.0]: \\
    $AE$ & 0.5 & 0.5 & 0.0 \\ \cline{3-4}
    & {[1.0,2.0],} & [1.0,8.0]: & [5.0,7.0], \\
    & {[4.0,5.0]:} &  & [2.0,4.0]: \\
    & 0.5 & 0.0 & 0.5 \\ \cline{3-4}
  \end{tabular}
\end{table}

Simplifying the intervals in Table \ref{fig:distance:matrixAeB} gives Table \ref{fig:distance:matrixAeBSimp}, resulting in
\begin{eqnarray*}
m_{D(AE,B)}&=&[1.0,8.0]:0.5,,\\
&&[2.0,4.0],[5.0,7.0]:0.5,\\
D(AE,B)&=&\{0.5\mid [1.0,2.0], 1.0 \mid [2.0,4.0],\\
&&0.5\mid [4.0,5.0], 1.0 \mid [5.0,7.0],\\
&&0.5 \mid [7.0,8.0]\}
\end{eqnarray*}

\begin{figure}[!htbp]
  \begin{center}
    \includegraphics[scale=0.23]{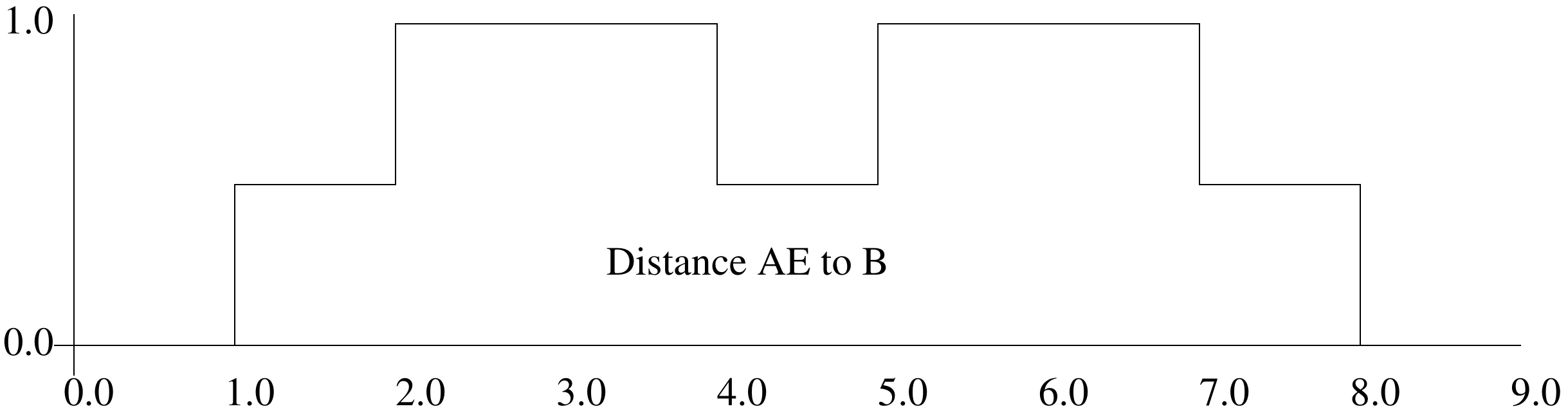}
    \caption{The multimodal FS of the distance between $AE$ and $B$.\label{fig:aetob}}
  \end{center}
\end{figure}

The resulting distance measure FS between $AE$ and $B$ is shown in Fig. \ref{fig:aetob}.

\subsection{Distances between non-normal multimodal FSs\label{sec:multimodal:non_normal}}
\begin{figure}[!htbp]
  \begin{center}
    \includegraphics[scale=0.23]{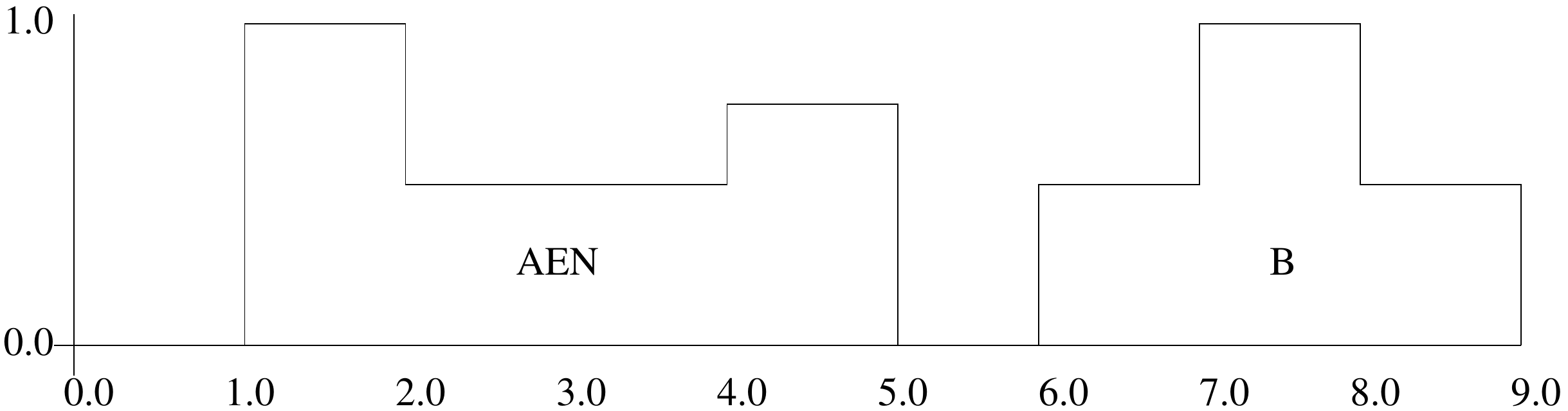}
    \caption{The multimodal FSs $AEN$ and $B$.\label{fig:aenandb}}
  \end{center}
\end{figure}

This section describes the effect of calculating the distance between a bimodal FS and a unimodal FS, where one of the bimodal modes is not normal. Take the two FSs $AEN$ and $B$, shown in Fig. \ref{fig:aenandb}. This has the slices at different levels and the diagonal rule cannot be directly applied to arrive at the assignment required. Slices must be taken at the same level. Table \ref{tab:distance:matrixAenBslices} rectifies this and the assignment is now down the diagonal.

\begin{eqnarray*}
m_{AEN} &=& [1.0,5.0]:0.5, [1.0,2.0],[4.0,5.0]:0.25,\\
&&[1.0,2.0]:0.25\\
\end{eqnarray*}

\begin{table}[!htbp]
  \setlength{\tabcolsep}{7pt}
  \caption{Example maximal distance measure between FSs $AEN$ and $B$.\label{tab:distance:matrixAenBslices}}
  \begin{tabular}{ c c|c|c|c|}
    & \multicolumn{1}{c}{} & \multicolumn{3}{c}{$B$}\\
    & \multicolumn{1}{c}{D($AEN$, $B$)} & \multicolumn{1}{c}{[6.0,9.0]:} & \multicolumn{1}{c}{[7.0,8.0]:}& \multicolumn{1}{c}{[7.0,8.0]:} \\
    & \multicolumn{1}{c}{} & \multicolumn{1}{c}{0.5} & \multicolumn{1}{c}{0.25} & \multicolumn{1}{c}{0.25} \\ \cline{3-5}
    & [1.0,5.0]: & [1.0,8.0]: & [2.0,7.0]: & [2.0,7.0]: \\
    & 0.5 & 0.5 & 0.0 & 0.0 \\ \cline{3-5}
    & {[1.0,2.0],} & [4.0,8.0], & [5.0,7.0], & [5.0,7.0],\\
    $AEN$ & {[4.0,5.0]:} & [1.0,5.0]: & [2.0,4.0]: & [2.0,4.0]: \\
    & 0.25 & 0.0 & 0.25 & 0.0 \\ \cline{3-5}
    & {[1.0,2.0],} & [4.0,8.0]: & [5.0,7.0]: & [5.0,7.0]:\\
    & 0.25 & 0.0 & 0.0 & 0.25 \\ \cline{3-5}
  \end{tabular}
\end{table}

Resulting in
{\arraycolsep=1pt\def\arraystretch{1}
\begin{eqnarray*}
m_{D(AEN,B)}&= &[1.0,8.0]:0.5,\\
&&[2.0,4.0],[5.0,7.0]:0.25, \\
&& [5.0,7.0]:0.25 \\
D(AEN,B)&=&\{0.5\mid [1.0,2.0], 0.75 \mid [2.0,4.0],\\
&&0.5\mid [4.0,5.0],1.0 \mid [5.0,7.0],\\
&&0.5 \mid [7.0,8.0]\}
\end{eqnarray*}
}

\begin{figure}[!htbp]
  \begin{center}
    \includegraphics[scale=0.23]{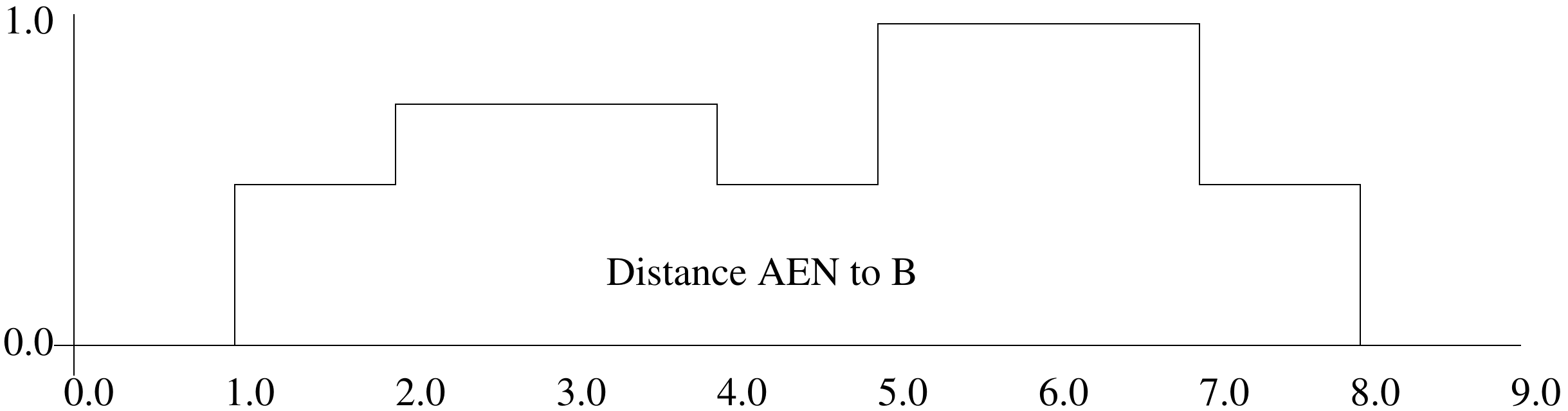}
    \caption{The multimodal FS showing the maximally calculated distance between $AEN$, $AM$ extended and not normalised, and $B$.\label{fig:distanceaenandbmax}}
  \end{center}
\end{figure}

The maximal distance measure between AEN and B, see Table \ref{tab:distance:matrixAenBslices}, results in a bimodal FS, see Fig. \ref{fig:distanceaenandbmax}, with one mode lower than the other.

\enlargethispage{-0.4cm}
\section{Conclusions\label{sec:conclusions}}
This paper has introduced MA based distance measures extending the work reported in \cite{McCulloch:2013:DDM}. The distance results are FSs and calculating the maximum likelihood values from the sets indicates that the measures accord with intuition, and is a better result than the centre of gravity approach. Ignoring the type 2 restrictions is an assumption that is likely to be broken often, however the result is computable directly and is more general and easier than the multiplicative method. The number of orthogonal assignments rises with the increased precision of the FS leading an assignment down the diagonal being a less restrictive assignment but is a useful compromise. 

Demonstrations have shown the effects with both normal and non-normal as well as convex and non-convex FSs, and though the paper has dealt with very blocky FSs which simplifies the calculations, the work generalises to countably continuous FSs.

\bibliographystyle{IEEEtran}
\bibliography{../bib/corpus}
\end{document}